\documentclass[journal]{IEEEtran}
\ifCLASSINFOpdf
\else
\fi

\usepackage{ulem}
\usepackage{times}
\usepackage{epsfig}
\usepackage{graphicx}
\usepackage{amsmath}
\usepackage{amssymb}
\usepackage{amsmath}
\DeclareMathOperator*{\argmin}{argmin}
\usepackage[pagebackref=true,breaklinks=true,letterpaper=true,colorlinks,bookmarks=false]{hyperref}

\begin{document}
%
\title{Unsupervised Learning of Global Registration of Temporal Sequence of Point Clouds}
%
%
%

\author{Lingjing Wang*, Yi Shi*, Xiang Li, Yi Fang
\thanks{* equal contribution.}
\thanks{L.Wang is with MMVC Lab, the Department of Electrical Engineering, New York University Abu Dhabi, UAE, e-mail: lingjing.wang@nyu.edu. X.Li is with the MMVC Lab, New York University Abu Dhabi UAE, e-mail: xl1845@nyu.edu. Y.Shi is with the MMVC Lab, New York University, New York, NY11201, USA, e-mail: ys3237@nyu.edu. Y.Fang is with MMVC Lab, Dept. of ECE, NYU Abu Dhabi, UAE and Dept. of ECE, NYU Tandon School of Engineering, USA, e-mail: yfang@nyu.edu.}
\thanks{Corresponding author: Yi Fang. Email: yfang@nyu.edu}}

%
%

\markboth{IEEE ROBOTICS AND AUTOMATION LETTERS,~Vol.~14, No.~8, August~2015}%
{Shell \MakeLowercase{\textit{et al.}}: Bare Demo of IEEEtran.cls for IEEE Journals}
%



\maketitle

\begin{abstract}
Global registration of point clouds aims to find an optimal alignment of a sequence of 2D or 3D point sets. In this paper, we present a novel method that takes advantage of current deep learning techniques for unsupervised learning of global registration from a temporal sequence of point clouds. Our key novelty is that we introduce a deep Spatio-Temporal REPresentation (STREP) feature, which describes the geometric essence of both temporal and spatial relationship of the sequence of point clouds acquired with sensors in an unknown environment. In contrast to the previous practice that treats each time step (pair-wise registration) individually, our unsupervised model starts with optimizing a sequence of latent STREP feature, which is then decoded to a temporally and spatially continuous sequence of geometric transformations to globally align multiple point clouds. We have evaluated our proposed approach over both simulated 2D and real 3D datasets and the experimental results demonstrate that our method can beat other techniques by taking into account the temporal information in deep feature learning. 
\end{abstract}


%
\IEEEpeerreviewmaketitle

\section{Introduction}
Registration on point sets is defined as finding the point-wise correspondence, either rigid or non-rigid transformation, that can optimally transform the source point set to the target one. Global registration of three-dimensional point cloud data is to find spatial geometric transformations between pairs in a sequence of local point cloud observations. All local observations can then be aligned into a single shape and brought into the same global coordinate. The applications of global registration are related to a wide range of industrial applications such as autonomous driving, medical imaging, and large-scale 3D reconstruction \cite{myronenko2009image,klaus2006segment,maintz1998survey,besl1992method,raguram2008comparative,yuille1988computational}. Given practical sensors have a limitation on detection angle or solid obstacles, the acquisition of multiple partially-overlapped point cloud scans from different positions and camera poses is inevitable. A global registration process has to be performed to get an accurate global scene. The key challenge behind this task is the extraction and comparison of geometric information among a series of local scenes.  \\

\begin{figure}
\begin{center}
\includegraphics[height=3.2cm]{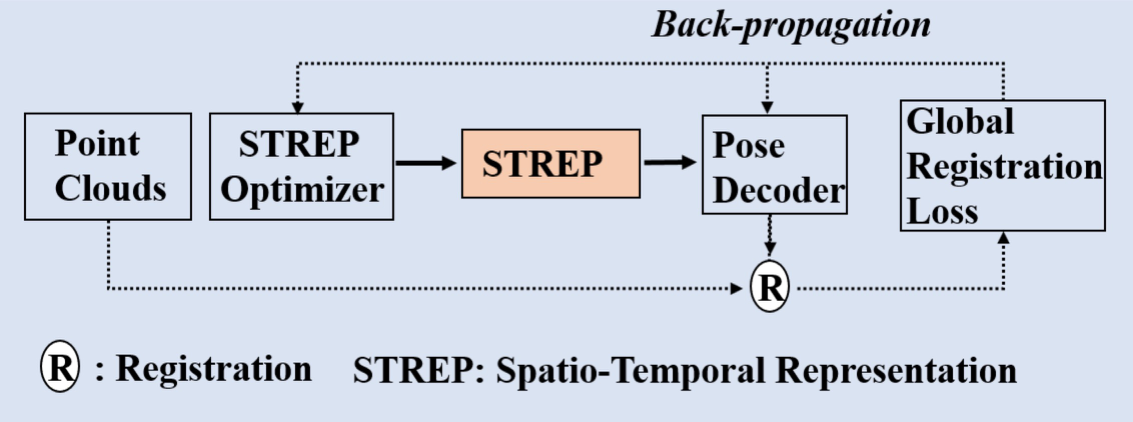}
\end{center}
\caption{The general pipeline of our proposed deep learning-based method. It starts with optimizing a randomly initialized latent spatio-temporal representation (STREP) feature, which links consecutive frames as a whole. Then STREP is further then decoded to a camera pose to locate the local point set in a global frame.}
\label{first}
\end{figure}
The classical methods such as Iterative closest point (ICP)\cite{besl1992method} and its variants \cite{yang2015go,zhou2016fast} usually approach this problem in an optimization process to iteratively minimize a pre-defined alignment loss between the transformed source point sets and their corresponding target ones to reach the optimal set of parameters of a geometric transformation. 

\begin{figure*}
\begin{center}
\includegraphics[height=9cm]{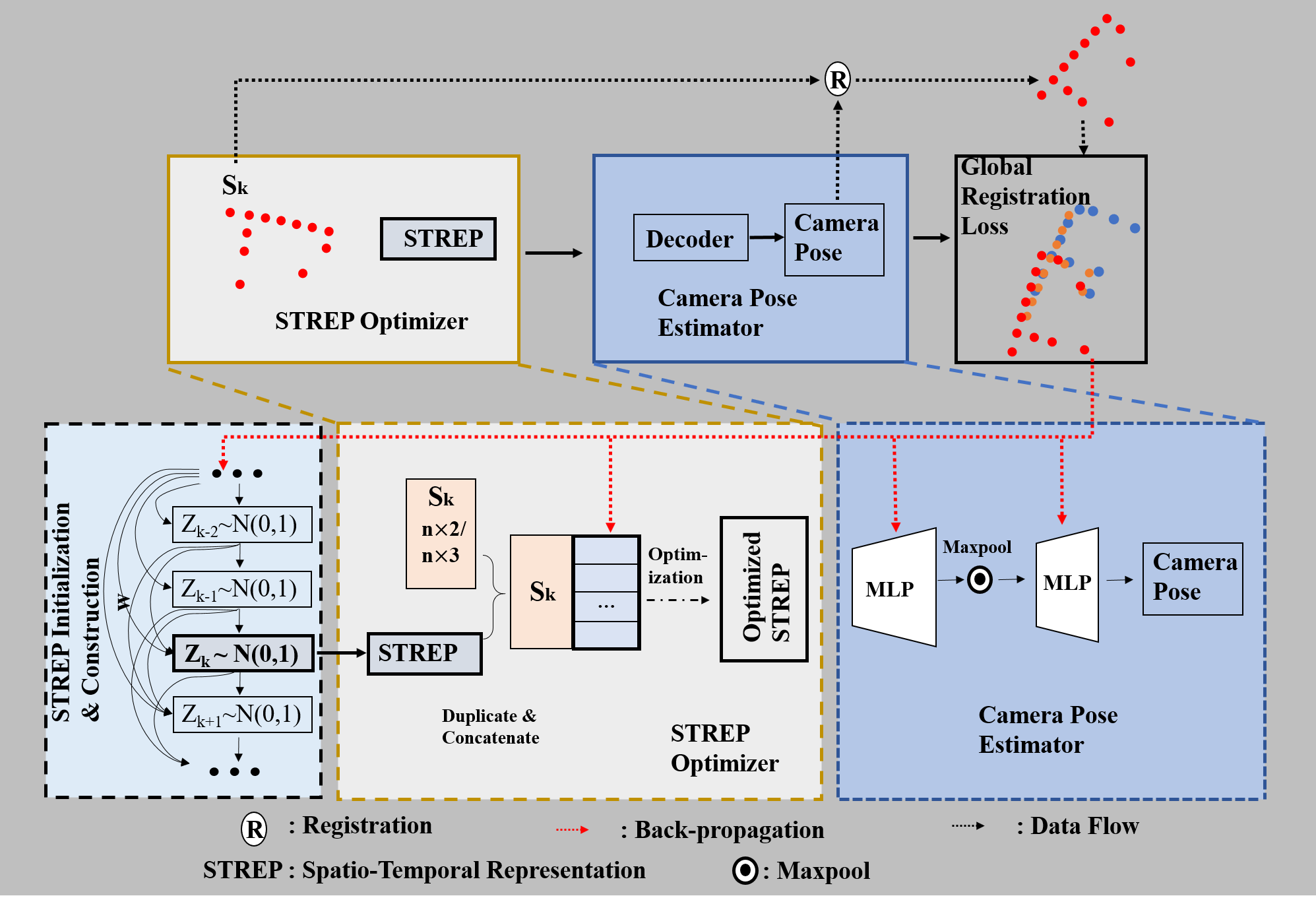}
\end{center}
\caption{Our pipeline. For a pair of input source and target point sets, instead of learning a correlation tensor between them, we concatenate an initial randomized vector z to each point of source shape and then design a decoder network to regress the translation and rotation matrix. The alignment loss between the transformed source point set and target point set will be back-propagated both on $\theta$ and initialized vector z.}
\label{main}
\end{figure*}

Deep learning models have recently achieved a great success in learning of point registrations \cite{Lu2019DeepVCPAE,deepicp,Wang_2019_NeurIPS}. However, there are some challenges for this specific learning task. Firstly, considering that the sequential registration task can be decomposed into a series of pairwise registration tasks, common questions such as how to extract features from 3D irregular non-grid point sets as well as how to explicitly model the geometric relationship between two point sets are still challenging for learning-based methods. Secondly, as a sequential registration problem, errors in registration of one pair may hurt the registration performance of the following pairs. Without an elegant solution to deal with this problem, the registration performance as a whole would be limited. For example, a recent learning-based research \cite{Ding_2019_CVPR} suffers from those issues, their deep networks only extract spatial information which undermines the overall performance. Therefore, how to deal with the sequential impact during the optimization process is another main challenging obstacle here. 

In this research, we propose an approach to consider both local spatial correlation and longer-term temporal sequential information sharing. Our method works in an unsupervised optimization manner where no extra training data and labels are required and the model focuses on aligning the given observations in series of local point sets that were scanned in different positions in a given scene. As shown in \ref{main}, each observation (i.e frame) has a corresponding learnable latent vector z as a temporal memory component with the purpose of conveying correlation information from its previous observations to the alignment process of other point sets. We proposed an encoder-free (i.e. no need to define an explicit feature encoding model) learning architecture where the latent vectors are optimized to attached to point sets and then sent into decoders to regress  camera poses. During the optimization process, the pre-defined loss can be back-propagated directly to all the latent vectors. Our network design avoids the design of an explicit encoder networks for extracting the latent feature representation vectors for 3D irregular non-grid point cloud. In addition, the new design of combination of the optimization and learning enables further fine-tuning on the unseen data with the guidance of a unsupervised alignment loss for model's better generalization ability. In contrast, the conventional neural network based models do not have the flexibility in fine-tuning at testing phase. Furthermore, in order to control the depth of temporal memory, each latent vector is further defined as the recurrent weighted sum of previous latent vector controlled by a weight factor w, as shown in the left-hand side of Figure \ref{main}. This mechanism enables global temporal information sharing which mitigates the sequential impact of pair-wise error mentioned earlier. 

The contributions of our work are three-fold: 1) We introduce an encoder-free temporal-spatial correlation latent vector that enforces a better alignment of sequential observations. Our design does not require a specific encoder model to extract the feature that represents the relationship between a pair of irregular non-grid point clouds.
2) Based on our proposed temporal latent vector, we introduce a latent fusion mechanism for accessing information of past frames in the sequence using our aggregated latent vectors.
3) We achieved a state-of-the-art experiment results on registration of both 2D and 3D datasets in which our model outperforms other approaches on solving scenes with complicated details.

\section{Related Works}
\subsection{Pairwise registration}
\noindent \textbf{Iterative methods.} The development of optimization algorithms to estimate the rigid or non-rigid geometric transformations in an iterative routine has been attracting attention in decades. The Iterative Closest Point (ICP) algorithm \cite{besl1992method} is one successful solution for rigid registration. It initializes an estimation of a rigid function and then iteratively chooses corresponding points to refine the transformation. However, the ICP algorithm is reported to be vulnerable to the selection of corresponding points for initial transformation estimation. Go-ICP \cite{yang2015go} was further proposed by Yang et al. to leverage BnB scheme for searching the entire 3D motion space to solve the local initialization problem brought by ICP. Zhou et al. proposed Fast Global Registration \cite{zhou2016fast} for registration of partially overlapping 3D surfaces. TPS-RSM algorithm was proposed by Chui and Rangarajan \cite{chui2000new} to estimate parameters of non-rigid transformation with a penalization on second-order derivatives. As a classical non-parametric method, coherence point drift (CPD) was proposed by Myronenko et al. \cite{myronenko2007non}, which successfully introduced a process of fitting Gaussian mixture likelihood to align the source point set with the target one. With the penalization term on the velocity field, the algorithm enforces the motion of the source point set to be coherent during the registration process. The existing classical algorithms achieved great success for the registration task. Even though all these methods state the registration task as an independent optimization process for every single pair of source and target point sets, they greatly inspire us for designing our learning-based system.\\ 

\noindent \textbf{Learning-based methods.} In recent years, learning-based methods achieved great success in many fields of computer vision \cite{qi2017pointnet,verma2018feastnet,zeng20173dmatch,wang20173densinet,wang2018unsupervised,wang2020few}. Especially recent works started a trend of directly learning geometric features from cloud points (especially 3D points), which motivates us to approach the point set registration problem using deep neural networks \cite{rocco2017convolutional,balakrishnan2018unsupervised,zeng20173dmatch,qi2017pointnet,verma2018feastnet,masci2015geodesic,li2019pc,wang2020unsupervised,wang2019non}. PointNetLK \cite{aoki2019pointnetlk} is proposed by Aoki et al to leverage the newly proposed PointNet algorithm for directly extracting features from Point cloud with the classical Lucas $\&$ Kanade algorithm for rigid registration of 3D point sets. Liu et al. proposed FlowNet3D \cite{liu2019flownet3d} to treat 3D point cloud registration as a motion process between points. Similarly, FlowNet3D leverages PointNet structure to extract the correlation relationship between source and target point sets firstly to further regress on the sense flow. \cite{robert17}\cite{dai2017b} proposes removing fused frames when they are gravely inconsistent during dealing with sequence data. Recently Wang et al. proposed Deep Closest Point \cite{wang2019deep} which firstly leverages DGCNN structure to extract the features from point sets and then regress the desired transformation based on it. All these methods have achieved outstanding registration performance. Moreover, in related fields, Balakrishnan et al. \cite{balakrishnan2018unsupervised} proposed a voxelMorph CNN architecture to learn the registration field to align two volumetric medical images. For registration of 2D images \cite{chen2019arbicon,rocco2017convolutional}, an outstanding registration model was proposed by Rocco et al. \cite{rocco2017convolutional}. This work mainly focuses on a parametric approach for 2D image registration. The parameters of both rigid and non-rigid function (TPS) can be predicted by a CNN-based structure from learning the correlation relationship between a pair of source and target 2D images. For learning-based registration solutions listed above, there is a major challenge about how to effectively model the relationship between the source and target object in a learning-based approach. For example, \cite{rocco2017convolutional} proposed a correlation tensor between the feature maps of source and target images. \cite{balakrishnan2018unsupervised} leverages a U-Net based structure to concatenate features of source and target voxels. \cite{liu2019flownet3d} 
\cite{aoki2019pointnetlk} uses PointNet based structure and \cite{wang2019deep} uses DGCNN structure to learn the features from a point set for further registration. However, designing a model to extract the features from point sets is troublesome and it is even more challenging to define pair-wise correlation relationship. In this paper, we propose a encoder-free structure to skip this encoding step. We initialize a random latent vector without pre-defining a model, which is to be optimized with the weights of network from the loss back-propagation process.
\subsection{Global registration}
In comparison to pairwise registration, sequential/global point clouds registration is much less explored. There are several previous research \cite{evangelidis2014generative,choi2015robust,torsello2011multiview} tried to solve related tasks. For example, one of them tried to register a sequence of point clouds one by one into a global frame. This framework does not change the existing registration result after adding a new point cloud, which accumulates errors during the process. A global cost function over a graph of sensor poses was proposed by \cite{choi2015robust,theiler2015globally} for predicting the drifts. The most recent research \cite{Ding_2019_CVPR} is one of the most successful DNN based approaches for sequential point set registration tasks. It proposed an elegant two-step approach, solving the transformation matrix and then estimating the global mapping. The whole optimization process is unsupervised and each camera pose is regressed from the input local captured point sets.

\section{Methods}
We introduce our approach in the following sections. First, we define the learning-based global registration problem in section 3.1. In section 3.2, we introduce the spatial-temporal feature learning module. We also explain the purpose and mechanism of temporal latent fusion. Section 3.3 illustrates our loss components. In section 3.4, we discuss the detail of the model optimization process. 

\subsection{Problem Statement}
For a given sequence of k input point sets $\bold{S} = \{ S_i\}_{i=1,...,k}$, where $S_i \subset \mathbb{R}^N$ (N=2 or N=3). We define the learning-based optimization task for global registration in the following way. We assume the existence of a parametric function $g_{\theta}(S_i) = \phi_i$ using a neural network structure, where $\phi_i$ is a set of the parameters of sensor pose including the location coordinates and rotation angle ($\phi_i$ has a size of 4/6 depending on 2D/3D data). $\theta$ represents all of the parameters in the deep neural networks. Each input local point set can be then transformed by the calculated rotation and translation operations defined by camera pose parameters $\phi$ according to a global frame. For a given training dataset $\bold{D}$, we can define a sequential registration loss by comparing transformed neighboring local input point sets. A global registration loss can be further defined on the correctness of combined transformed local frames and the global ground truth. Stochastic gradient descent can be used for weights $\theta$'s optimization towards minimizing our loss function:
\begin{equation}
\begin{split}
\bold{\theta^{optimal}} =\argmin_{\theta, \gamma}[\mathbb{E}_{\{S_i\}\sim \bold{D}}[\mathcal{L_{\gamma}}(g_{\theta}(\bold{S_1}),...,g_{\theta}(\bold{S_k})],
\end{split}
\end{equation}
where $\mathcal{L_\gamma}$ represents a global registration loss. $\mathcal{L_\gamma}$ can be an unsupervised learnable objective function with weights $\gamma$. After training, a trained model is able to perform an inference to obtain the parameters $\phi$ based on the optimized optimal $\theta^{optimal}$ in the neural network structure.  

\subsection{Spatial-Temporal feature learning.}

For a sequence of input point sets $\bold{S}=\{S_i\}_{i=1,2,...,k}$, learning shape's feature is our first task. A better feature learning architecture directly has an impact on the performance of camera pose regression for each local shape. Firstly, we discuss the vanilla network without using a temporal latent component. As the backbone of the model, a PointNet-based structure is adopted for points' features learning. More specifically, assuming there are n points in each point set, a multi-layer perceptron (MLP) is utilized for learning the spatial feature. The architecture includes multiple MLP layers with ReLu activation function: $\{g_i\}_{i=1,2,...,s}$, such that: $g_i : \mathbb{R}^{v_{i}}\to \mathbb{R}^{v_{i+1}}$, where $v_{i}$ and $v_{i+1}$ are the dimension of input layer and output layer. A maxpool layer is used to extract the global spatial feature, defined as $\mathbf{L^{spatial}}$. Therefore, $\forall \bold{S_k}$,
\begin{equation}
\begin{split}
\bold{L_k^ {spatial}}=Maxpool\{g_sg_{s-1}...g_1(\bold{x_i})\}_{\bold{x_i}\in \bold{S_k}}
\end{split}
\end{equation}

However, this only extracts the spatial information from each local observation. During our experiments, we discover that only utilizing spatial feature is rather inept at solving frames with complicated details where the effort made in aligning one pair of neighboring observations sometimes impact the alignment of others. Moreover, if a sequence of local point sets has to be broken up into several batches because of the restriction of computational resources, alignment between different batches is prone to failure for the lack of loss constraints. To address this issue, we introduce our solution to take shareable temporal information into account during the feature extraction phase. The following structure aims to learn the temporal-spatial feature, defined as $\mathbf{L^{spatial-temporal}}$, for each local input point set. 

We initialize a set of trainable latent vectors $\bold{\tilde{Z}} = \{ \tilde{z_i} \}_{i=1,2,...,k}$. For each $S_i$, we have $\tilde{z_i} \sim \mathcal{N}(0,1)$ and $\tilde{z_i} \in \mathbb{R}^b$. The latent vectors represent a frame-wise shareable sequential memory. By conditioning model loss on the latent vectors in an optimization manner, it allows more concrete information sharing between nearby observations after several iterations. Furthermore, we propose a latent fusion method where the past latent information helps the alignment of the present pair of observations. A memory decay weight $w \in \mathbb{R^b}$ is also defined so that past frame information can be utilized for the alignment task at hand. Empirically, it is designed as a learnable vector.

The latent vectors $\bold{Z}$ can be expressed as: 
  \[
    \left\{
                \begin{array}{ll}
                  z_1=\tilde{z_1}\\
                  z_k=\tilde{z_k}+w z_{k-1} \text{ for } k>1\\
                \end{array}
              \right.
  \]

By leveraging the same structure of formula (2), with the extra help of the extra temporal feature, the model is guided to view multiple pair-wise tasks in a more global manner that greatly improves the performance. $\bold{L_k^{spatial-temporal}}$ is defined as :
\begin{equation}
\begin{split}
\bold{L_k^{spatial-temporal}}=Maxpool\{g_sg_{s-1}...g_1([\bold{x_i},z_k])\}_{\bold{x_i}\in \bold{S_k}}
\end{split}
\end{equation}
where the notation [*,*] represents the concatenation of vectors in the same domain. 
Using the learned feature from previous part as the input, $\forall \bold{L_k}$, a encoding network f composed with MLP and ReLU layers is proposed.
\begin{equation}
\begin{split}
\bold{\mathbf{\phi_k}}=f(\mathbf{L_k})
\end{split}
\end{equation}
  $\phi_k$ is the desired camera pose for shape $\mathbf{S_k}$. 

After the pose is obtained through optimization, we can transform the local point sets $\mathbf{S_k}$ into a global version $\mathbf{G_k}$. The multiple point set frames can then be stacked together and get the final global scene $\mathbf{Gm}$.
\subsection {Loss Components.}
Our loss components are composed of two parts: pair-wise registration loss and global registration loss.

For pair-wise registration, a typical loss component used in researches\cite{Lu2019DeepVCPAE} is variations of squared Euclidean Distance that measures the distance between points in the target (ground truth) observation and points in the transformed source observation. It can be expressed by:
\begin{equation}\label{Z_op}
\begin{split}
\bold{\mathcal{L}}(A,B) =\sum^{N}_{i=0}||a_i-b_i||^2
\end{split}
\end{equation} 

The Chamfer Distance is a less strict criterion that measures the average distance between points in one observation and their closest neighbors located in the other point set observation.  This criterion describes how similar a pair of observations are without the constraint of knowing the point-wise matching. Since we do not presume the direct point correspondence, it is chosen for further experiment. The Chamfer distance between point sets A and B can be defined as:

\begin{equation} 
\begin{split}d_{\text{Chamfer}}(A,B)
 &= \sum_{a\in A}\min_{b\in B}||a-b||^2_2\\
 &+ \sum_{b\in B}\min_{a\in A}||a-b||^2_2
\end{split}
\end{equation}

The total registration loss $\mathcal {L}_{ch}$ can be defined on point set $\mathbf{S_i}$ with its neighbors $\mathbf{G_j}$ as:
\begin{equation} 
\mathcal{L}_{local}= \sum_{i=1}^{K} \sum_{j \in \mathcal{N}(i)}d(\mathbf{S_i}, \mathbf{S_j})
\end{equation}

In \cite{Ding_2019_CVPR}, a global loss component is utilized for their mapping task so that a smoother and more accurate global scene can be generated. In essence, it is a binary cross entropy loss for evaluating global frame accuracy. It estimates the occupancy status of the coordinates in the aligned global frame. For the coordinate of any actual point belong to the ground truth global frame, it should be marked as occupied. Any point between the scanning sensor and any ground truth point should be labeled as unoccupied. A sampling process will be performed to gather coordinates of unoccupied points. For a aligned local observation $\mathbf{F_j} \subset \mathbf{Fm}$; $\mathbf{BE}$[*] as binary cross entropy loss and $\mathbf{S}$ [*] as unoccupied point sampling procedure.
\begin{equation} 
\mathcal{L}_{global}= \frac{1}{K}\sum_{j=1}^{K}BE[\mathbf{F_j},1]+BE[S(\mathbf{F_j}),0]
\label{map}
\end{equation}
As mentioned, in scenes with complicated yet isometric details, an alignment of pair-wise observations sometimes leads to local optimums with regard to our global registration task. The error it caused will have a sequential impact on the registration of later pairs. Therefore, we consider \ref{map} as a global registration loss to be experimented further. 

Experimented with multiple loss function settings, the summation of a pair-wise Chamfer Loss and a global registration loss described by \ref{map} yields the best results.

\subsection{Optimization process.}

Learning features from point clouds is a challenging task, but it would be more challenging if we need to model the temporal relationship among local patches in a neighborhood. We propose a model-free structure for learning the latent vector and thus skip all the subjective designs as in previous researches. As discussed in section 3.2, we leverage a trainable randomly initialized latent vectors $z$ to realize temporal-spatial feature learning. Instead of extracting the latent vector from inputs using an encoder structure, our latent vector z is initialized from a random vector, which is sampled from a Gaussian Distribution $\mathcal{N}(0, 1)$. The whole training phase is an optimization process on the given observations where both the weights of the decoder and the latent vectors are updated. The model is capable of aligning given observations of a scene in an ``unsupervised" optimization manner for this task, we do not use and additional information for our setting in comparison to other methods. It cannot perform any inference operation on new data. 

For a given series of observations $\mathbf{D}$, we use stochastic gradient descent-based algorithm to optimize not only parameter set $\theta$ as the weights of the model and the parameter set $\gamma$ in mapping loss, but also the set of latent vectors $\mathbf{z} = \{z_1,...,z_n\}$ and the temporal weights $w$ for minimizing the expected loss function:

\begin{equation}
\begin{split}
\bold{\theta^{optimal}, z^{optimal}, w^{optimal},\gamma^{optimal}} \\
=\argmin_{\theta, \mathbf{z}, w, \mathbf{\gamma}}[\mathbb{E}_{\{\bold{S_i}\}\sim \bold{D}}[\mathcal{L}_\gamma(g_{\theta}([\bold{S_1},\mathbf{z_1}])),...,g_{\theta}([\bold{S_k},\mathbf{z_k}])]],
\end{split}
\end{equation}
where $\mathcal{L_\gamma}$ represents a pre-defined loss. For the testing cases in dataset $\mathbf{T}$, we fix the optimized parameters $\tilde{\theta}=\theta^{optimal}$, $\tilde{\gamma} = \gamma^{optimal}$ in the network and loss function and the temporal weights $\tilde{w}=w^{optimal}$. We need to optimize the following function again:

\begin{equation}
\bold{z^{optimal}}=\argmin_{\mathbf{z}}[\mathbb{E}_{\{\bold{S_i}\}\sim \bold{T}}[\mathcal{L}_{\tilde{\gamma}}(g_{\tilde{\theta}}([\bold{S_1},\mathbf{z_1}])),...,g_{\tilde{\theta}}([\bold{S_k},\mathbf{z_k}])]],
\end{equation}
After this optimization process, the desired transformation is $\phi_i=g_{\tilde{\theta}}(\bold{S_i},\mathbf{z_i^{optimal}})$ and the transformed source shape $\mathbf{S_i'}=\phi_i(\mathbf{S_i})$

\begin{figure}[h]
\begin{center}
\includegraphics[height=12cm]{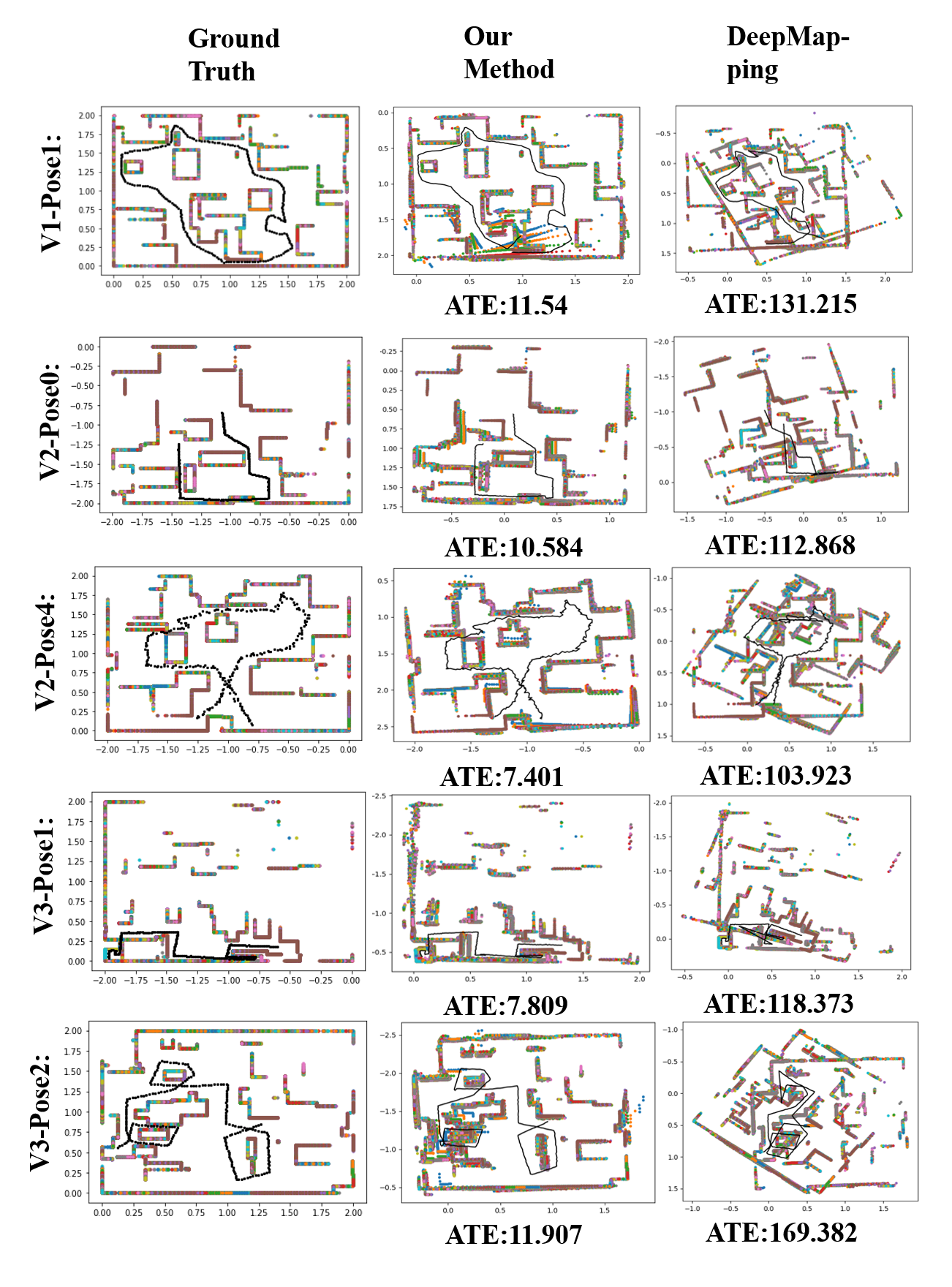}
\end{center}
\caption{Qualitative result of 2D synthesized images registration. Each color in the point clouds represent a local frame.}
\label{exp1}
\end{figure}

\section{Experiments}
In this section, we implement the following experiments to validate the performance of our proposed model for global point set registration. For a fair comparison, we follow the experiment setting from the state-of-the-art learning-based method \cite{Ding_2019_CVPR} on two datasets: a simulated 2D Lidar point cloud dataset and a real 3D dataset called Active Vision Dataset (AVD) \cite{ammirato2017dataset}. In section 4.1, we test our model's performance on synthesized 2D simulated point clouds. Experiments on real-world 3D active vision dataset and comparison with state-of-the-art methods are discussed in section 4.2. Moreover, Section 4.3 discusses the ablation study for our proposed method. Our method is implemented with PyTorch 1.0. Adam optimizer is our default optimization algorithm and we set a learning rate of 0.001 for network and 0.0001 for the temporal latent vector z's. We use two metrics: the absolute trajectory error (ATE) and the point distance (We use point-wise L2 distance in this paper) between a ground truth point cloud and a registered one for quantitative comparison. 

\begin{figure*}
\begin{center}
\includegraphics[width=13.5cm]{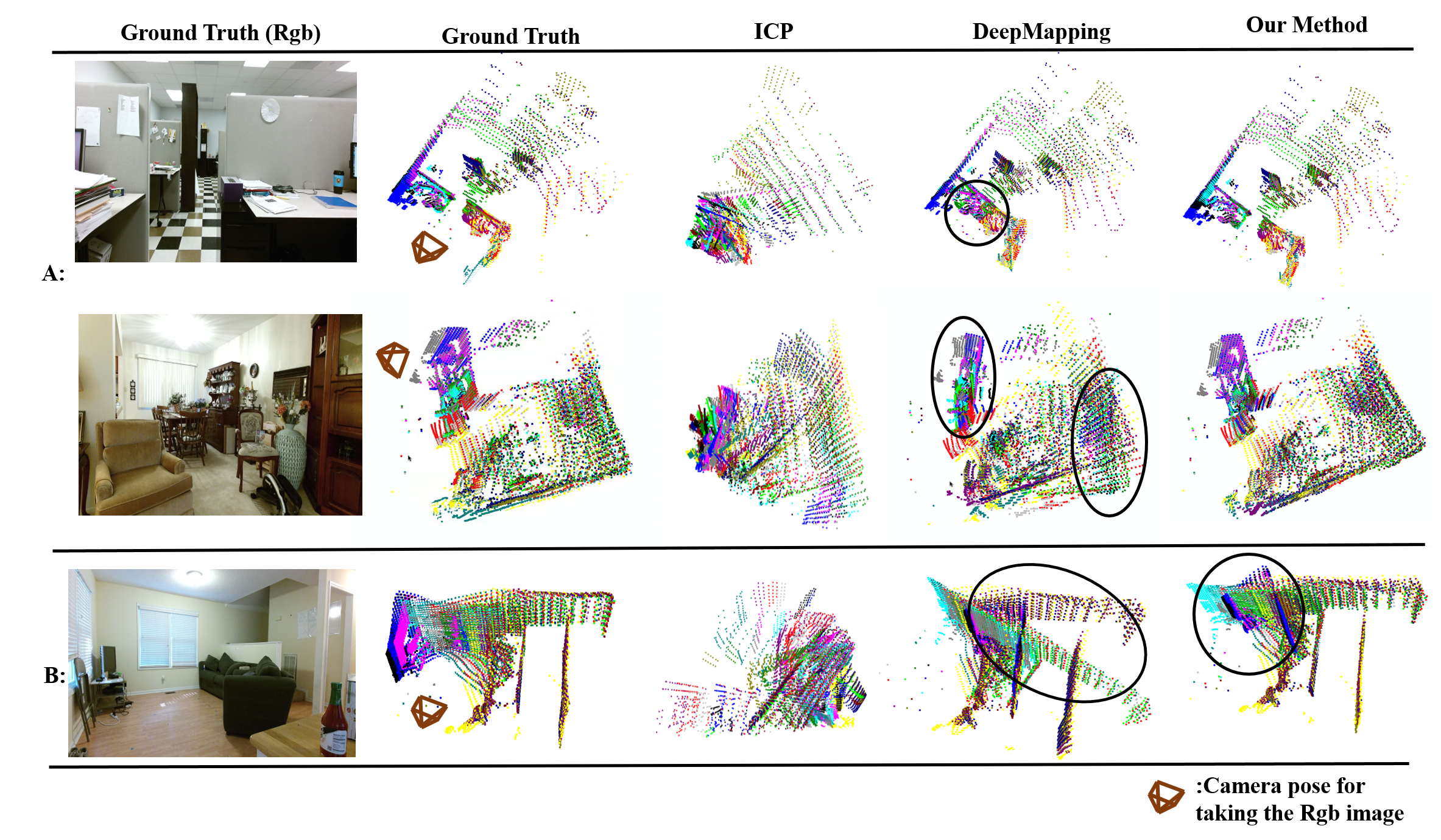}
\end{center}
\caption{Qualitative result of 3D AVD registration. Due to the high complexity of the indoor environment, we provide the 2D RGB picture which is captured at the small camera position for reference. Each color in the point clouds represents a local frame. We mark the area for your attention in black circles.}
\label{exp2}
\end{figure*}

\subsection{Experiments on 2D Simulated Point Cloud}

\noindent \textbf{Dataset: }
We use 20 trajectories from three global environments that are provided by authors of \cite{Ding_2019_CVPR}. More specifically, assuming that we have a large 2D binary image as the global ground truth environment, we need to sample the trajectory as a moving agent to obtain local frames. The trajectory includes a sequence of 2D coordinates which represents the pose location. Between consecutive scans, a rotation from -10 degree to 10 degree and a translation of 0-16 pixels are randomly generated. The sampled 2D translation vector and rotation angle is then our ground truth local camera pose. \\

\noindent \textbf{Baseline: }
We use \cite{Ding_2019_CVPR} as our baseline model since it is the current state-of-the-art result, which has significantly better performance than more traditional methods such as ICP, GoICP, PSO, Direct Optimization on the same task. \\

\noindent \textbf{Implementation: }
Hereby we introduced our network architecture that was illustrated in Fig \ref{main}. The temporal latent vectors are firstly attached to each point and sent into the MLP to calculate a camera pose. The series of points sets to be aligned are fed to the model together. This part of the network is composed of:  C[64]-C[256]-C[1024]-M[1024]-F[512]-F[128]-F[3]. C[*] represents 1D convolution with kernel size 3. M[*] represents 1D max-pooling layer. F[*] denotes fully-connected layers. Then the poses can be utilized to transform local observations to a global one and further conditioned by the difference in combined local observations that have been aligned and the ground truth. The network for computing global registration is composed of: F[64]-F[256]-F[512]-F[256]-F[128]-F[1]. For our temporal latent vector z, we set its dimension to 16. During the training process, both latent vectors and weights in the  network are optimized. We notice that the whole learning process is unsupervised. We set the batch size to 128. \\

\noindent \textbf{Result: } For most 2D cases, during our experiments, we notice that our method and the baseline method can achieve similar results. Since the 2D synthesized cases are simple and only contain horizontal and vertical lines, we pick some failed cases obtained by the baseline method caused by their lack of temporal information sharing for comparison. We do not notice any case that the baseline method succeeds in registration while our model fails. The qualitative results are presented in Figure \ref{exp1} and for each case, we demonstrate the ATE below the qualitative result. As shown in Figure \ref{exp1}, our method can succeed in the registration of all these five cases when the baseline method fails. The average ATE of our model for such cases is around 10 but the average ATE of the baseline method for these cases is more than 100. \\

\begin{figure}
\begin{center}
\includegraphics[width=7.5cm]{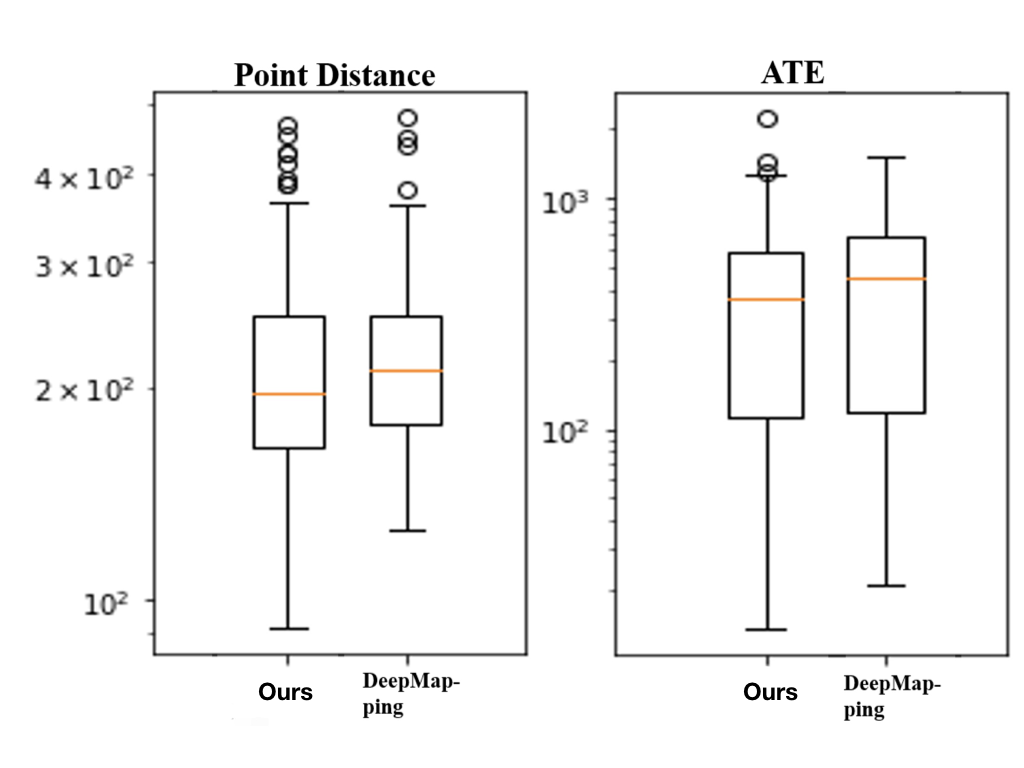}
\end{center}
\caption{Quantitative result of 3D AVD images registration.}
\label{stat2}
\end{figure}
\begin{figure*}
\begin{center}
\includegraphics[width=14.5cm]{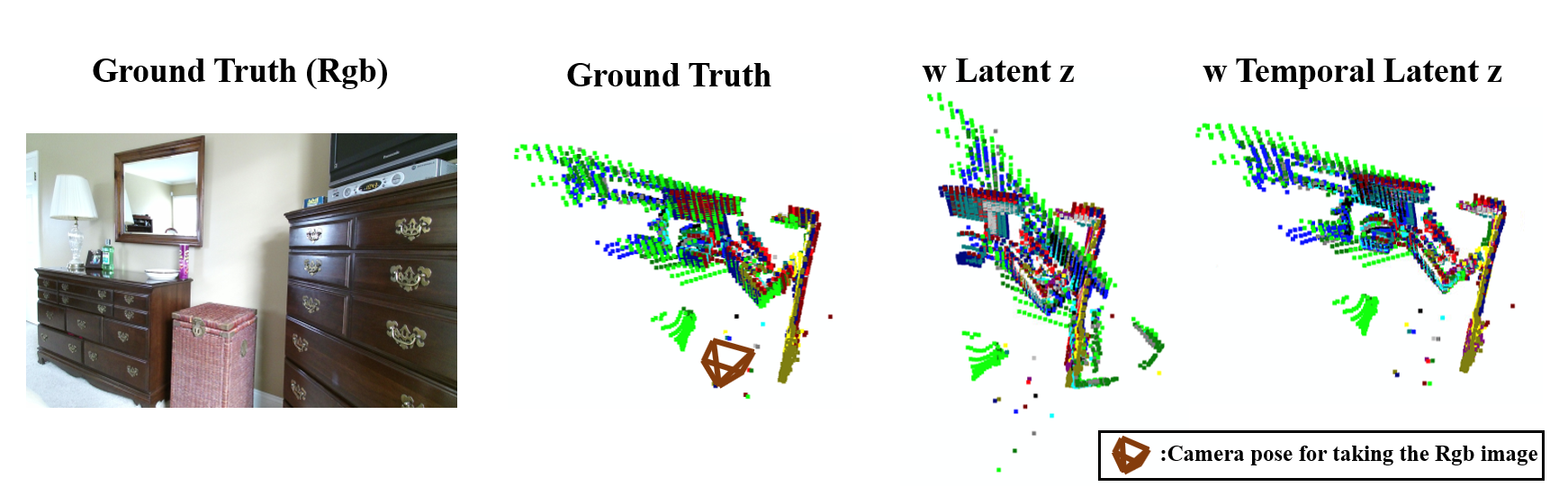}
\end{center}
\caption{Qualitative result of ablation study for cases with temporal weights versus w/o temporal weights. Due to the high complexity of the indoor environment, we provide the 2D RGB picture which is captured at the small camera position for reference.}
\label{exp3}
\end{figure*}

\begin{figure}[h]
\begin{center}
\includegraphics[width=7.5cm]{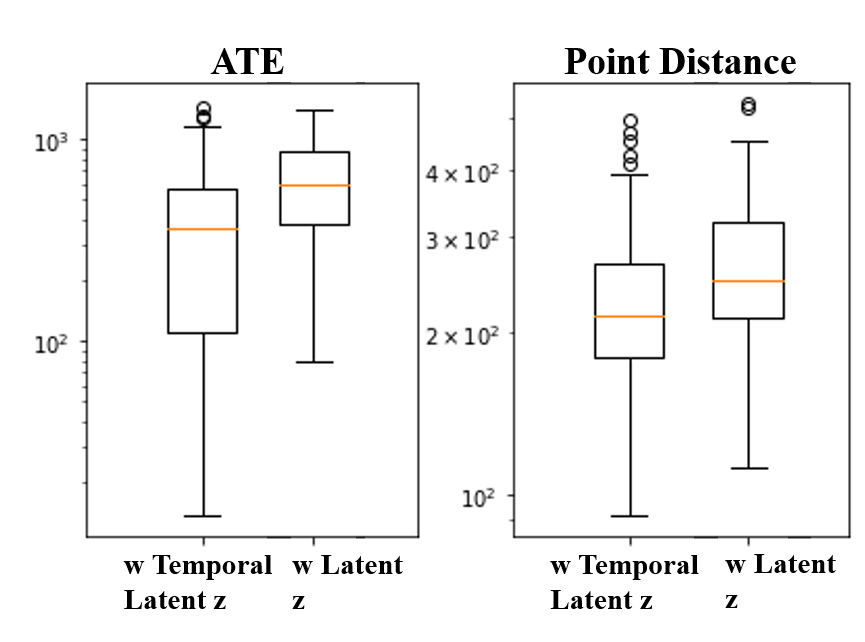}
\end{center}
\caption{Quantitative result of ablation study for cases with temporal weights versus w/o temporal weights. }
\label{stat3}
\end{figure}
\subsection{Experiments on Active Vision Dataset}

\noindent \textbf{Dataset:}
we further test our method on a real 3D indoor dataset: Active Vision Dataset \cite{ammirato2017dataset}. This dataset provides indoor scans with RGB-D datasets. AVD dataset has a set of discrete visiting points on a rectangular grid with a fixed width of 300mm. 12 views of images are taken at the position of each point by rotating the camera 30 degrees. We only use depth images and transform them into point clouds. 100 trajectories are randomly collected from the dataset using our custom script. Each trajectory contains 16 frames. During the generation of our custom trajectories and sequential point cloud observations, we emphasize the variety of the indoor scenes where all types of scenes (bedroom, office, living room) are presented and the trajectories are sampled evenly on each scene. We will provide our script and generated trajectories online.   \\

\noindent \textbf{Baseline: } We do not use scores reported in any previous research since they did not publish their generated trajectories or any relative generation code. Both models are evaluated on our own generated trajectories for a fair comparison. The results are thus different from the original paper.\\

\noindent \textbf{Implementation: }  For our temporal latent vectors z, we increase its dimension to 24 in hope of caring more information. During the training process, both latent vectors and weights in the network are optimized. We decrease the batch size to 8 for the great memory consumption caused by a large 3D scene. \\

\noindent\textbf{Result: } The selected qualitative results are presented in Figure \ref{exp2} and the quantitative results are presented in Figure \ref{stat2}. For quantitative result shown in Figure \ref{stat2}, we notice that both our average registration performance and the lower bound for the 100 trajectories are lower than the baseline. Bar charts are provided to illustrate that the comparison in Figure \ref{exp2} is not handpicked. Our method is better at dealing with frames with complicated details as shown in Figure \ref{exp2}. The RGB images are captured at the location marked by a small black cross as the viewport. In Row A, we notice that the baseline model mistakenly registers the door area and the left wall in comparison to our result in the first case. In the second case, due to the complexity of the furniture's arrangement in the right lower corner, the baseline model failed to align them correctly while ours achieves much better performance. In Row B, both our model and the baseline model failed to perfectly align each local frame. Our model failed to align the window area but the mismatching area achieved by the baseline model is much broader than ours.  

\subsection{Ablation study}
\noindent \textbf{Dataset: } We sample extra trajectories of 20 from AVD dataset as explained in previous section.\\

\noindent \textbf{Baseline: } For verifying the advantages of temporal latent vector space. We remove the links between latent vectors for each scene but we still concatenate a latent vector z on each point of the input shape as our baseline model. We compare it with ours to further verify our temporal structure's efficiency.\\

\noindent \textbf{Implementation: }  For the baseline model, there is no additional initialized weight to link neighbor latent vectors.  We further verify that the temporal setting of our model mainly contributes to the final result instead of purely searching for the latent vector space. \\

\noindent \textbf{Result: }One selected qualitative result is presented in Figure \ref{exp3} and the quantitative statistics are presented in Figure \ref{stat3}. As shown in Figure \ref{exp3}, we can clearly see the baseline model failed to register the case but our method can achieve much better performance. For the quantitative result presented in Figure \ref{stat3}, our model achieves better results than the baseline model for both metrics, which further confirms the contributions of our temporal settings with regard to the performance enhancement. 

\section{Conclusion And Discussion} A Spatial-Temporal latent learning mechanism is introduced for sequential point set registration problems. It provides three unique advantages for global registration:1) the features that represent irregular non-grid point clouds can be learned without the design of specific 3D feature encoders, 2) it improves global registration by considering the spatial smoothness and temporal consistency in point sets, and 3) it enhances the learning of global registration for point clouds acquired in unseen scenes which empowers our unsupervised optimization framework. Through a great number of experiments on 2D/3D point cloud data, our method is proved to be more effective when compared to the current state-of-the-art method. 

\ifCLASSOPTIONcaptionsoff
  \newpage
\fi



%

\bibliographystyle{IEEEtran}
\bibliography{egbib}

\end{document}